%% file: main.tex
\documentclass[10pt,twocolumn,letterpaper]{article}

\usepackage{iccv2023AuthorKit/iccv}
\usepackage{times}
\usepackage{epsfig}
\usepackage{graphicx}
\usepackage{amsmath}
\usepackage{amssymb}
\usepackage[accsupp]{axessibility} %

\include{custom_includes.tex}

\usepackage{rotating}
\usepackage{multirow}
\usepackage{booktabs}
\usepackage{caption}
\usepackage{color}
\usepackage{makecell}
\usepackage{bbm}
\usepackage{balance}
\usepackage[font=small]{caption}
\usepackage{pdfpages}

\usepackage{graphicx}
\usepackage{amsmath}
\usepackage{amssymb}
\usepackage{booktabs}
\usepackage{multirow}
\usepackage{booktabs}
\usepackage{algorithm}
\usepackage{algorithmicx}
\usepackage{mathtools}
\usepackage{adjustbox}
\usepackage{import}
\usepackage[noend]{algpseudocode}
\usepackage{caption}
\usepackage{subcaption}

\newlength\savewidth

\renewcommand{\paragraph}[1]{\vspace{1.25mm}\noindent\textbf{#1}}

\newcommand{\app}{\raise.17ex\hbox{$\scriptstyle\sim$}}
\usepackage[pagebackref=false, breaklinks=true, letterpaper=true, colorlinks, urlcolor=gray,
citecolor=citecolor, linkcolor=linkcolor, bookmarks=false]{hyperref}
\usepackage{cleveref}

\definecolor{citecolor}{HTML}{0071BC}
\definecolor{linkcolor}{HTML}{ED1C24}

\iccvfinalcopy %

\def\iccvPaperID{4482} %

\ificcvfinal\fi

\begin{document}

\title{Humans in 4D: Reconstructing and Tracking Humans with Transformers}

\author{Shubham Goel\hspace{0.75em} 
Georgios Pavlakos\hspace{0.75em} 
Jathushan Rajasegaran\hspace{0.75em} 
Angjoo Kanazawa$^*$\hspace{0.75em} 
Jitendra Malik$^*$\\
{\tt\small \{shubham-goel, pavlakos, jathushan, kanazawa\}@berkeley.edu, malik@eecs.berkeley.edu}\\
University of California, Berkeley\\
}

\twocolumn[{%
\renewcommand\twocolumn[1][]{#1}%
\maketitle
\begin{center}
    \vspace{-0.26in}
    \centerline{
   \includegraphics[width=0.98\textwidth,clip]{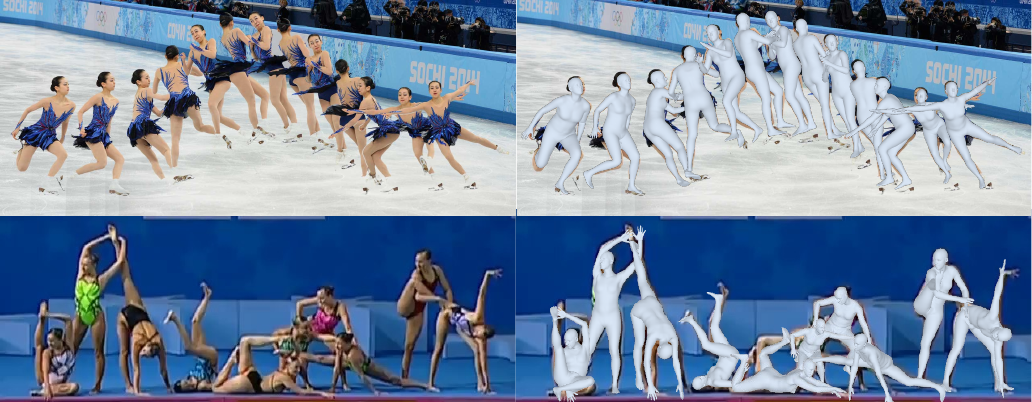}
     }
    \vspace{-0.05in}
   \captionof{figure}{{\bf A ``transformerized'' view of Human Mesh Recovery}.
   We describe \approachName, a fully transformer-based approach for 3D human pose and shape reconstruction from a single image.
   Besides impressive performance across a wide variety of poses and viewpoints, HMR 2.0 also acts as the backbone of an improved system for jointly reconstructing and tracking Humans in 4D (\systemName).
   Here, we see output reconstructions from \approachName~for each 2D detection in the left image.
   }
   \vspace{-0.02in}
\label{fig:teaser}
\end{center}%
}]

\ificcvfinal\thispagestyle{empty}\fi
\ificcvfinal\thispagestyle{empty}\fi

\begin{abstract}
We present an approach to reconstruct humans and track them over time.
At the core of our approach, we propose a fully ``transformerized'' version of a network for human mesh recovery.
This network, \approachName, advances the state of the art and shows the capability to analyze unusual poses that have in the past been difficult to reconstruct from single images.
To analyze video, we use 3D reconstructions from \approachName~as input to a tracking system that operates in 3D. 
This enables us to deal with multiple people and maintain identities through occlusion events.
Our complete approach, \systemName, achieves state-of-the-art results for tracking people from monocular video. 
Furthermore, we demonstrate the effectiveness of \approachName~on the downstream task of action recognition, achieving significant improvements over previous pose-based action recognition approaches.
Our code and models are available on the project website:
\url{https://shubham-goel.github.io/4dhumans/}.
\end{abstract}

\section{Introduction}
In this paper, we present a fully transformer-based approach for recovering 3D meshes of human bodies from single images, and tracking them over time in video.  We obtain unprecedented accuracy in our single-image 3D reconstructions (see Figure~\ref{fig:teaser}) even for unusual poses where previous approaches struggle. In video, we link these reconstructions over time by 3D tracking, in the process bridging gaps due to occlusion or detection failures. 
These 4D reconstructions can be seen on the project webpage.

Our problem formulation and approach can be conceived as the ``transformerization'' of previous work on human mesh recovery, HMR~\cite{kanazawa2018end} and 3D tracking, PHALP~\cite{rajasegaran2022tracking}. 
Since the pioneering ViT paper \cite{dosovitskiy2020vit},
the process of ``transformerization'', \ie, converting models from CNNs or LSTMs to transformer backbones, has advanced rapidly across multiple computer vision tasks, \eg, \cite{carion2020end,fan2021multiscale,he2022masked,li2022exploring,peebles2022scalable,wu2023multiview}. Specifically for 2D pose (2D body keypoints) this has already been done by ViTPose~\cite{xu2022vitpose}. We take that as a starting point and we develop a new version of HMR, which we call \approachName~to acknowledge its antecedent.

We use \approachName~to build a system that can simultaneously reconstruct and track humans from videos.
We rely on the recent 3D tracking system, PHALP~\cite{rajasegaran2022tracking}, which we simplify and improve using our pose recovery.
This system can reconstruct Humans in 4D, which gives the name to our method, \systemName.
\systemName~can be deployed on any video and can jointly track and reconstruct people in video.
The functionality of creating a tracking entity for every person is fundamental towards analyzing and understanding humans in video.
Besides achieving state-of-the-art results for tracking on the PoseTrack dataset~\cite{andriluka2018posetrack}, we also apply \approachName~ on the downstream application of action recognition.
We follow the system design of recent work,~\cite{rajasegaran2023action}, and we show that the use of \approachName~can achieve impressive improvements upon the state of the art on action recognition on the AVA v2.2 dataset.

This paper is unabashedly a systems paper. We make design choices that lead to the best systems for 3D human reconstruction and tracking in the wild. Our model is publicly available on the project webpage. There is an emerging trend, in computer vision as in natural language processing, of large pretrained models which find widespread downstream applications and thus justify the scaling effort. \approachName~is such a large pre-trained model which could potentially be useful not just in computer vision, but also in robotics~\cite{patel2022learning,peng2018sfv,vasilopoulos2020reactive}, computer graphics~\cite{weng2022humannerf}, bio-mechanics~\cite{pearl2023fusion}, and other fields where analysis of the human figure and its movement from images or videos is needed.

Our contributions can be summarized as follows:
\begin{enumerate}
    \item We propose an end-to-end ``transformerized'' architecture for human mesh recovery, \approachName. Without relying on domain-specific designs, we outperform existing approaches for 3D body pose reconstruction.
    \item Building on \approachName, we design \systemName~that can jointly reconstruct and track humans in video, achieving state-of-the-art results for tracking.
    
    \item We show that better 3D poses from \approachName~result in better performance on the downstream task of action recognition, finally contributing to the state-of-the-art result (42.3 mAP) on the AVA benchmark.
\end{enumerate}

\section{Related Work}

\noindent
{\bf Human Mesh Recovery from a Single Image.}
Although, there have been many approaches that estimate 3D human pose and shape relying on iterative optimization, \eg, SMPLify~\cite{bogo2016keep} and variants~\cite{guan2009estimating,lassner2017unite,pavlakos2019expressive,rempe2021humor,tiwari2022pose,zanfir2018monocular}, for this analysis we will focus on approaches that directly regress the body shape from a single image input.
In this case, the canonical example is HMR~\cite{kanazawa2018end}, which uses a CNN to regress SMPL~\cite{loper2015smpl} parameters.
Since its introduction, many improvements have been proposed for the original method.
Notably, many works have proposed alternative methods for pseudo-ground truth generation, including using temporal information~\cite{arnab2019exploiting},
multiple views~\cite{leroy2020smply},
or iterative optimization~\cite{kolotouros2019learning,joo2021exemplar,pavlakos2022human}.
SPIN~\cite{kolotouros2019learning} proposed an in-the-loop optimization that incorporated SMPLify~\cite{bogo2016keep} in the HMR training.
Here, we also rely on pseudo-ground truth fits for training, and we use~\cite{kolotouros2021probabilistic} for the offline fitting.

More recently, there have been works that propose more specialized designs for the HMR architecture.
\mbox{PyMAF}~\cite{zhang2021pymaf,zhang2022pymaf} incorporates a mesh alignment module for the regression of the SMPL parameters.
PARE~\cite{kocabas2021pare} proposes a body-part-guided attention mechanism for better occlusion handling.
HKMR~\cite{georgakis2020hierarchical} performs a prediction that is informed by the known hierarchical structure of SMPL.
HoloPose~\cite{guler2019holopose} proposes a pooling strategy that follows the 2D locations of each body joints.
Instead, we follow a design without any domain-specific decisions and we show that it outperforms all previous approaches.

Many related approaches are making non-parametric predictions, \ie, instead of estimating the parameters of the SMPL model, they explicitly regress the vertices of the mesh.
GraphCMR~\cite{kolotouros2019convolutional} uses a graph neural network for the prediction,
METRO~\cite{lin2021end} and FastMETRO~\cite{cho2022cross} use a transformer,
while Mesh Graphormer~\cite{lin2021mesh} adopts a hybrid between the two.
Since we regress the SMPL model parameters, instead of the locations of mesh vertices, we are not directly comparable to these. However, we show how we can use a fully ``transformerized'' design for HMR.

\noindent
{\bf Human Mesh \& Motion Recovery from Video.}
To extend Human Mesh Recovery over time, most methods use the basic backbone of HMR~\cite{kanazawa2018end} and propose designs for the temporal encoder that fuses the per-frame features.
HMMR~\cite{kanazawa2019learning} uses a convolutional encoder on features extracted from HMR~\cite{kanazawa2018end}.
VIBE~\cite{kocabas2020vibe}, MEVA~\cite{luo20203d} and TCMR~\cite{choi2021beyond} use a recurrent temporal encoder.
DSD~\cite{sun2019human} combines convolutional and self-attention layers, while MAED~\cite{wan2021encoder} and t-HMMR~\cite{pavlakos2022human} employ a transformer-based temporal encoder.
Baradel~\etal~\cite{baradel2021leveraging, baradel2022posebert} also used a transformer for temporal pose prediction, while operating directly on SMPL poses.
One key limitation of these approaches is that they often operate in scenarios where tracking is simple~\cite{kanazawa2019learning, zhang2019predicting}, \eg, videos with a single person or minimal occlusions.
In contrast to that, our complete \systemName~approach is also solving the tracking problem.

\begin{figure*}
    \centering
    \small
    \includegraphics[width=\linewidth]{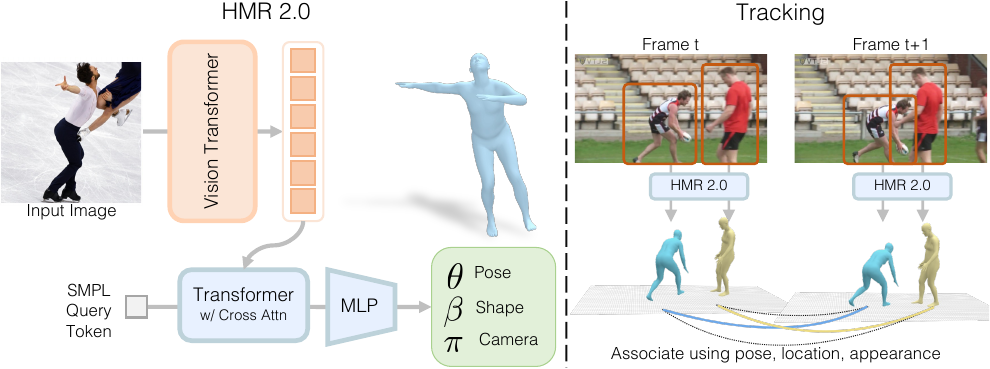}
    \caption{{\bf Overview of our approach.} Left: \approachName~is a fully ``transformerized'' version of a network for Human Mesh Recovery.
    Right: We use \approachName~as the backbone of our \systemName~system, that builds on PHALP \cite{rajasegaran2022tracking}, to jointly reconstruct and track humans in 4D.}
    \label{fig:approach}
\end{figure*}

\noindent
{\bf Tracking People in Video.}
Recently, there have been approaches that demonstrate state-of-the-art performance for tracking by relying on 3D human reconstruction from HMR models, \ie, T3DP~\cite{rajasegaran2021tracking} and PHALP~\cite{rajasegaran2022tracking}.
In these methods, every person detection is lifted to 3D using an HMR network~\cite{pavlakos2022human} and then tracking is performed using the 3D representations from lifting~\cite{rajasegaran2021tracking} and prediction~\cite{rajasegaran2022tracking} to track people in video.
Empirical results show that PHALP works very well on multiple tracking benchmarks (the main requirement is that the images have enough spatial resolution to permit lifting of the people to 3D).
We use these tracking pipelines, and particularly PHALP, as a task to evaluate methods for human mesh recovery.

\noindent
{\bf Action Recognition.}
Action recognition is typically performed using appearance features from raw video input.
Canonical examples in this category include SlowFast~\cite{feichtenhofer2019slowfast} and MViT~\cite{fan2021multiscale}.
Simultaneously, there are approaches that use features extracted from body pose information, \eg, PoTion~\cite{choutas2018potion} and JMRN~\cite{shah2022pose}.
A recent approach, LART~\cite{rajasegaran2023action}, demonstrates state-of-the-art performance for action recognition by fusing video-based features with features from 3D human pose estimates.
We use the pipeline of this approach and employ action recognition as a downstream task to evaluate human mesh recovery methods.

\section{Reconstructing People}

\subsection{Preliminaries}

\noindent
{\bf Body Model.} 
The SMPL model \cite{SMPL:2015} is a low-dimensional parametric model of the human body. Given input parameters for pose ($\theta \in \real^{24\times3\times3}$) and shape ($\beta \in \real^{10}$), it outputs a mesh $M \in \real^{3\times N}$ with $N=6890$ vertices. The body joints $X\in\real^{3\times k}$ are defined as a linear combination of the vertices and can be computed as $X=MW$ with fixed weights $W \in \real^{N\times k}$. Note that pose parameters $\theta$ include the body pose parameters $\theta_{b} \in \real^{23\times3\times3}$ and the global orientation $\theta_{g} \in \real^{3\times3}$.

\noindent
{\bf Camera.}   
We use a perspective camera model with fixed focal length and intrinsics $K$. Each camera $\pi = (R,t)$ consists of a global orientation $R \in \real^{3\times 3}$ and translation $t \in \real^3$. Given these parameters, points in the SMPL space (\eg, joints $X$) can be projected to the image as $x = \pi(X) = \Pi(K(RX + t))$, where $\Pi$ is a perspective projection with camera intrinsics $K$. Since $\theta$ already includes a global orientation, in practice we assume $R$ as identity and only predict camera translation $t$.

\noindent
{\bf HMR.}
The goal of the human mesh reconstruction (HMR) task is to learn a predictor $f(I)$ that given a single image I, reconstructs the person in the image by predicting their 3D pose and shape parameters. Following the typical parametric approaches \cite{kanazawa2018end, kolotouros2019learning}, we model $f$ to predict $\Theta = [\theta, \beta, \pi] = f(I)$ where $\theta$ and $\beta$ are the SMPL pose and shape parameters and $\pi$ is the camera translation.

\subsection{Architecture}
\label{sub:arch}

We re-imagine HMR \cite{kanazawa2018end} as an end-to-end transformer architecture that uses no domain specific design choices. 
Yet, it outperforms all existing approaches that have heavily customized architectures and elaborate design decisions. As shown in \Cref{fig:approach}, we use (i) a ViT \cite{dosovitskiy2020vit} to extract image tokens, and (ii) a standard transformer decoder that cross-attends to image tokens to output $\Theta$.

\noindent
{\bf ViT.} The Vision Transformer, or ViT~\cite{dosovitskiy2020vit} is a transformer~\cite{vaswani2017attention} that has been modified to operate on an image. 
The input image is first patchified into input tokens and passed through the transformer to get output tokens.
The output tokens are then passed to the transformer decoder. We use a ViT-H/16, the ``Huge'' variant with $16\times16$ input patch size. Please see SupMat for more details.

\noindent
{\bf Transformer decoder.} We use a standard transformer decoder~\cite{vaswani2017attention} with multi-head self-attention. It processes a single (zero) input token by cross-attending to the output image tokens and ends with a linear readout of $\Theta$.
We follow~\cite{kolotouros2019learning} and regress 3D rotations using the representation of~\cite{zhou2019continuity}.

\subsection{Losses}
Following best practices in the HMR literature \cite{kanazawa2018end, kolotouros2019learning}, we train our predictor $f$ with a combination of 2D losses, 3D losses, and a discriminator. Since we train with a mixture of datasets, each having different kinds of annotations, we employ a subset of these losses for each image in a mini-batch. We use the same losses even with pseudo-ground truth annotations. Given an input image $I$, the model predicts $\Theta = [\theta, \beta, \pi] = f(I)$. Whenever we have access to the ground-truth SMPL pose parameters $\theta^*$ and shape parameters $\beta^*$, we bootstrap the model predictions using an MSE loss: 
$$\loss_\texttt{smpl} = ||\theta - \theta^*||_2^2 + ||\beta - \beta^*||_2^2.$$ 
When the image has accurate ground-truth 3D keypoint annotations $X^*$, we additionally supervise the predicted 3D keypoints $X$ with an L1 loss: 
$$\loss_\texttt{kp3D} = ||X - X^*||_1.$$
When the image has 2D keypoints annotations $x^*$, we supervise projections of predicted 3D keypoints $\pi(X)$ using an L1 loss: 
$$\loss_\texttt{kp2D} = ||\pi(X) - x^*||_1.$$
Furthermore, we want to ensure that our model predicts valid 3D poses and use the adversarial prior in HMR \cite{kanazawa2018end}. It factorizes the model parameters into: (i) body pose parameters $\theta_b$, (ii) shape parameters $\beta$, and (iii) per-part relative rotations $\theta_i$, which is one 3D rotation for each of the 23 joints of the SMPL model. We train a discriminator $D_k$ for each factor of the body model, and the generator loss can be expressed as:
$$\loss_\texttt{adv} = \sum_k(D_k(\theta_b, \beta) - 1)^2.$$

\subsection{Pseudo-Ground Truth fitting}
We scale to unlabelled datasets (\ie, InstaVariety~\cite{kanazawa2019learning}, AVA~\cite{gu2018ava}, AI Challenger~\cite{wu2017ai}) by computing pseudo-ground truth annotations. Given any image, we first use an off-the-shelf detector~\cite{li2022exploring} and a body keypoints estimator~\cite{xu2022vitpose} to get bounding boxes and corresponding 2D keypoints. We then fit a SMPL mesh to these 2D keypoints using ProHMR~\cite{kolotouros2021probabilistic} to get pseudo-ground truth SMPL parameters $\theta^*$ and $\beta^*$ with camera $\pi^*$.

\begin{figure}[!h]
    \centering
    \small
    \includegraphics[width=\linewidth]{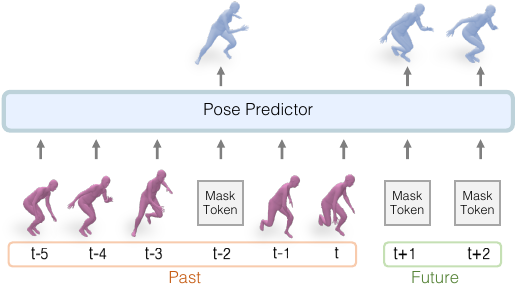}
    \caption{{\bf Pose prediction:} We train a BERT-style~\cite{devlin2018bert} transformer model on over 1 million tracks obtained from ~\cite{rajasegaran2023action}. This allow us to make future predictions and amodal completion of missing detections using the same model. To predict future poses ($t+1$, $t+2$, ...), we query the model with a mask-token using corresponding positional embeddings. Similarly for amodal completion, we replace missing detections with a masked token.}
    \label{fig:phalp_pose_transformer}
   \vspace{-1.8em}
\end{figure}

\section{Tracking People}

In videos with multiple people, we need the ability to associate people across time, \ie, perform tracking.
For this we build upon PHALP~\cite{rajasegaran2022tracking}, a state-of-the-art tracker based on features derived from HMR-style 3D reconstructions.
The basic idea is to detect people in individual frames, and ``lift'' them to 3D, extracting their 3D pose, location in 3D space (derived from the estimated camera), and 3D appearance (derived from the texture map). A tracklet representation is incrementally built up for each individual person over time. The recursion step is to predict for each tracklet, the pose, location and appearance of the person in the next frame, all in 3D, and then find best matches between these top-down predictions and the bottom-up detections of people in that frame after lifting them to 3D. The state represented by each tracklet is then updated by the incoming observation, and the process is iterated.
It is possible to track through occlusions because the 3D representation of a tracklet continues to be updated based on past history.

We believe that a robust pose predictor should also perform well, when evaluated on this downstream task of tracking, so we use the tracking metrics as a proxy to evaluate the quality of 3D reconstructions.
But first we needed to modify the PHALP framework to allow for fair comparison of different pose prediction models.
Originally, PHALP used pose features based on the last layer of the HMR network, \ie, a 2048-dimensional embedding space.
This limits the ability of PHALP to be used with different pose models (\eg, HMR 2.0, PARE, PyMAF etc.).
To create a more generic version of PHALP, we perform the modification of representing pose in terms of SMPL pose parameters, and we accordingly optimize the PHALP cost function to utilize the new pose distance.
Similarly, we adapt the pose predictor to operate on the space of SMPL parameters.
More specifically, we train a vanilla transformer model~\cite{vaswani2017attention} by masking random pose tokens as shown in the Fig~\ref{fig:phalp_pose_transformer}. 
This allows us to predict future poses in time, as well as amodal completion of missing detections.
With these modifications, we can plug in any mesh recovery methods and run them on any videos. We call this modified version PHALP$'$.

\paragraph{4DHumans.} Our final tracking system, \systemName, uses a sampling-based parameter-free appearance head and a new pose predictor (\Cref{fig:phalp_pose_transformer}). To model appearance, we texture visible points on the mesh by projecting them onto the input image and sampling color from the corresponding pixels. 

To track people in videos, previous approaches relied on off-the-shelf tracking approaches and used their output to reconstruct humans in videos (\eg, take the bounding boxes from tracking output and reconstruct people). For example, PHD~\cite{zhang2019predicting}, HMMR~\cite{kanazawa2019learning} can run on videos with only single person in the scene. In this work, we combine reconstruction and tracking into a single system and show that better pose reconstructions result in better tracking and this combined system can now run on any videos in the wild.

\section{Experiments}
In this section, we evaluate our reconstruction and tracking system qualitatively and quantitatively. First, we show that \approachName~outperforms previous methods on standard 2D and 3D pose accuracy metrics (\Cref{sub:pose_metrics}). Second, we show \systemName~is a versatile tracker, achieving state-of-the-art performance (\Cref{sub:tracking}).
Third, we further demonstrate the robustness and accuracy of our recovered poses via superior performance on the downstream application of action recognition (\Cref{sub:action}). Finally, we discuss the experimental investigation when designing \approachName~and ablate a series of design choices (\Cref{sub:design}).

\subsection{Setup}
\noindent
{\bf Datasets.} Following previous work, we use the typical datasets for training, \ie, Human3.6M~\cite{ionescu2013human3}, MPI-INF-3DHP~\cite{mehta2017monocular}, COCO~\cite{lin2014microsoft} and MPII~\cite{andriluka20142d}. Additionally, we use InstaVariety~\cite{kanazawa2019learning}, AVA~\cite{gu2018ava} and AI Challenger~\cite{wu2017ai} as extra data where we generate pseudo-ground truth fits.

\noindent
{\bf Baselines.} We report performance on benchmarks that we can compare with many previous works (\Cref{sub:pose_metrics}), but we also perform a more detailed comparison with recent state-of-the-art methods, \ie, PyMAF~\cite{zhang2021pymaf}, CLIFF \cite{li2022cliff}, HMAR~\cite{rajasegaran2022tracking}, PARE~\cite{kocabas2021pare}, and PyMAF-X~\cite{zhang2022pymaf}. For fairness, we only evaluate the body-only performance of PyMAF-X. %

\addtolength{\tabcolsep}{-5pt}
\begin{table}[t]
  \centering
  \footnotesize	
    \begin{tabular}{cl|cc|cc}
    \toprule
    \multicolumn{2}{c|}{\multirow{2}[4]{*}{Method}} & \multicolumn{2}{c|}{3DPW} & \multicolumn{2}{c}{Human3.6M} \\
\cmidrule{3-6}    \multicolumn{2}{c|}{} & MPJPE & PA-MPJPE & MPJPE & PA-MPJPE \\
    \midrule
    \multirow{5}[2]{*}{\begin{sideways}\rotatebox[origin=c]{0}{Temporal}\end{sideways}} & Kanazawa~\etal~\cite{kanazawa2019learning} & 116.5 & 72.6 & -     & 56.9 \\
          & Doersch~\etal~\cite{doersch2019sim2real} & -     & 74.7  & -     & - \\
          & Arnab~\etal~\cite{arnab2019exploiting} & -     & 72.2 & 77.8  & 54.3 \\
          & DSD~\cite{sun2019human} & -     & 69.5  & 59.1  & 42.4 \\
          & VIBE~\cite{kocabas2020vibe} & 93.5  & 56.5 & 65.9  & 41.5 \\
    \midrule
    \multirow{16}[4]{*}{\begin{sideways}\rotatebox[origin=c]{0}{Frame-based}\end{sideways}} & Pavlakos~\etal~\cite{pavlakos2018learning} & -     & -     & -     & 75.9 \\
          & HMR~\cite{kanazawa2018end} & 130.0   & 76.7 & 88.0    & 56.8 \\
          & NBF~\cite{omran2018neural} & -     &       & -     & 59.9 \\
          & GraphCMR~\cite{kolotouros2019convolutional} & -     & 70.2 & -     & 50.1 \\
          & HoloPose~\cite{guler2019holopose} & -     & -     & 60.3  & 46.5 \\
          & DenseRaC~\cite{xu2019denserac} & -     & -     & 76.8  & 48.0 \\
          & SPIN~\cite{kolotouros2019learning} & 96.9  & 59.2 & 62.5  & 41.1 \\
          & DecoMR~\cite{zeng20203d} & -     & 61.7$^\dagger$  & -     & 39.3$^\dagger$ \\
          & DaNet~\cite{zhang2019danet} &  -  &  56.9   & 61.5  & 48.6 \\
          & Song~\etal~\cite{song2020human} & -     &  55.9   & -  & 56.4 \\
          & I2L-MeshNet~\cite{moon2020i2l} & 100.0   & 60.0    & 55.7$^\dagger$  & 41.1$^\dagger$ \\
          & HKMR~\cite{georgakis2020hierarchical} & -     & -     & 59.6  & 43.2 \\
          & PyMAF~\cite{zhang2021pymaf}               & 92.8        & 58.9         & 57.7         & 40.5         \\
          & PARE~\cite{kocabas2021pare}                & 82.0& \tbest{50.9} & 76.8         & 50.6         \\
          & PyMAF-X~\cite{zhang2022pymaf}             & \sbest{78.0}& \sbest{47.1} & \tbest{54.2} & \tbest{37.2} \\
          & {\approachName}a      & \best{70.0}         & \best{44.5}  & \best{44.8}  & \sbest{33.6}  \\
          & {\approachName}b      & \tbest{81.3}        & 54.3         & \sbest{50.0}         & \best{32.4}         \\
    \bottomrule
    \end{tabular}%
    \caption{\textbf{Reconstructions evaluated in 3D:} Reconstruction errors (in mm) on the 3DPW and Human3.6M datasets. $^\dagger$ denotes the numbers evaluated on non-parametric results. Lower $\downarrow$ is better. Please see the text for details.
    }
  \label{tab:3d_metrics}%
\end{table}%
\addtolength{\tabcolsep}{5pt}

\addtolength{\tabcolsep}{-5pt}
\begin{table}[t]
  \centering
  \small
\begin{tabular}{@{}lccccccccccc@{}}
\toprule
\multicolumn{1}{c}{\multirow{2}{*}{Method}} & \multicolumn{2}{c}{LSP-Extended} &  & \multicolumn{2}{c}{COCO} &  & \multicolumn{2}{c}{PoseTrack} \\
\multicolumn{1}{c}{}                        & @0.05        & @0.1          &  & @0.05          & @0.1           &  & @0.05         & @0.1           \\ \midrule
PyMAF~\cite{zhang2021pymaf}                 & -            & -             &  & 0.68           & 0.86           &  & 0.77          & 0.92           \\
CLIFF~\cite{li2022cliff}                    & \tbest{0.32} & \tbest{0.66}  &  & 0.64           & 0.88           &  & 0.75          & 0.92           \\
PARE~\cite{kocabas2021pare}                 & {0.27}       & 0.60          &  & 0.72           & 0.91           &  & 0.79          & 0.93           \\
PyMAF-X~\cite{zhang2022pymaf}               & -            & -             &  & \sbest{0.79}   & \tbest{0.93}   &  & \tbest{0.85}  & \tbest{0.95}   \\
{\approachName}a                             & \sbest{0.38} & \sbest{0.72}  &  & \sbest{0.79}   & \sbest{0.95}   &  & \sbest{0.86}  & \sbest{0.97}   \\
{\approachName}b                             & \best{0.53}  & \best{0.82}   &  & \best{0.86}    & \best{0.96}    &  & \best{0.90}   & \best{0.98}   \\
\bottomrule
\end{tabular}
    \caption{\textbf{Reconstructions evaluated in 2D.} PCK scores of projected keypoints at different thresholds on the LSP-Extended, COCO, and PoseTrack datasets. Higher $\uparrow$ is better. 
    }
  \label{tab:2d_pck}%
\end{table}%
\addtolength{\tabcolsep}{5pt}

\subsection{Pose Accuracy}
\label{sub:pose_metrics}

\noindent
{\bf 3D Metrics.} 
For 3D pose accuracy, we follow the typical protocols of prior work, \eg,~\cite{kolotouros2019learning}, and we present results on the 3DPW test split and on the Human3.6M val split, reporting MPJPE, and PA-MPJPE in \Cref{tab:3d_metrics}.
Please notice that we only compare with methods that do not use the training set of 3DPW for training, similar to us.
We observe that with our {\approachName}a model, which trains only on the typical datasets, we can outperform all previous baselines across all metrics.
However, we believe that these benchmarks are very saturated and these smaller differences in pose metrics tend to not be very significant.
In fact, we observe that by a small compromise of the performance on 3DPW, our \mbox{{\approachName}b} model, which trains for longer on more data (AVA~\cite{gu2018ava}, AI Challenger~\cite{wu2017ai}, and InstaVariety~\cite{kanazawa2019learning}), achieves results that perform better on more unusual poses than what can be found in Human3.6M and 3DPW. We observe this qualitatively and from performance evaluated on 2D pose reprojection (\Cref{tab:2d_pck}). Furthermore, we observe that \mbox{{\approachName}b} is a more robust model 
and use it for evaluation in the rest of the paper.

\noindent
{\bf 2D Metrics.} 
We evaluate 2D image alignment of the generated poses by reporting PCK of reprojected keypoints at different thresholds on LSP-Extended~\cite{johnson2011learning}, COCO validation set~\cite{lin2014microsoft}, and Posetrack validation set~\cite{andriluka2018posetrack}. Since \mbox{PyMAF(-X)}~\cite{zhang2021pymaf,zhang2022pymaf} were trained using LSP-Extended, we do not report numbers for that part of the table.
Notice in \Cref{tab:2d_pck}, that {\approachName}b consistently outperforms all previous approaches. On LSP-Extended, which contains unusual poses, {\approachName}b~achieves PCK@0.05 of 0.53, which is $1.6\times$ better than the second best (CLIFF) with 0.32. For PCK@0.05 on easier datasets like COCO and PoseTrack with less extreme poses, {\approachName}b~still outperforms the second-best approaches but by narrower margins of 9\% and 6\% respectively. {\approachName}a also outperforms all baselines, but is worse than {\approachName}b, especially on harder poses in LSP-Extended.

\begin{figure*}
    \centering
    \includegraphics[width=0.99\textwidth]{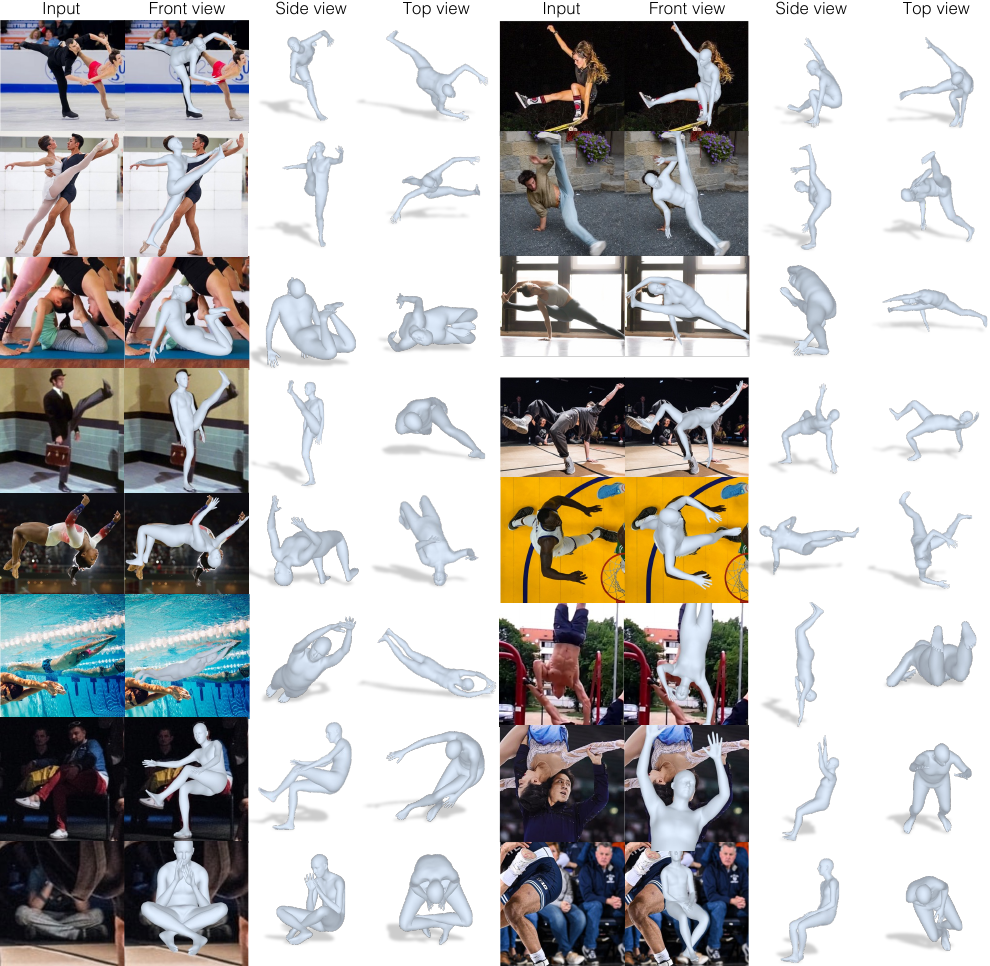}
    \vspace{+0.25em}
    \caption{{\bf Qualitative evaluation of \approachName.} 
    For each example we show: a) the input image, b) the reconstruction overlay, c) a side view, d) the top view.
    To demonstrate the robustness of \approachName, we visualize results for a variety of settings - for unusual poses (rows 1-4), for unusual viewpoints (row 5) and for images with poor visibility, extreme truncations and extreme occlusions (rows 6-8).}
    \label{fig:qual_big}
    \vspace{+0.25em}
\end{figure*}

\noindent
{\bf Qualitative Results.} We show qualitative results of \approachName~in \Cref{fig:qual_big}. We are robust to extreme poses and partial occlusions. Our reconstructions are well-aligned with the image and are valid when seen from a novel view.
Moreover, we compare with our closest competitors in \Cref{fig:pose-comparison}.
We observe that PyMAF-X and particularly PARE often struggle with more unusual poses, while \approachName~returns more faithful reconstructions.

\subsection{Tracking}
\label{sub:tracking}

For tracking, we first demonstrate the versatility of the modifications introduced by PHALP$'$, which allow us to evaluate 3D pose estimators on the downstream task of tracking. Then, we evaluate our complete system, \systemName, with respect to the state of the art.

\paragraph{Evaluation Setting.} Following previous work \cite{rajasegaran2021tracking, rajasegaran2022tracking}, we report results based on IDs (ID switches), MOTA~\cite{kasturi2008framework}, IDF1~\cite{ristani2016performance}, and HOTA~\cite{luiten2021hota} on the Posetrack validation set using the protocol of~\cite{rajasegaran2022tracking}, with detections from Mask R-CNN~\cite{he2017mask}.

\begin{figure*}
    \centering
    \includegraphics[width=\linewidth]{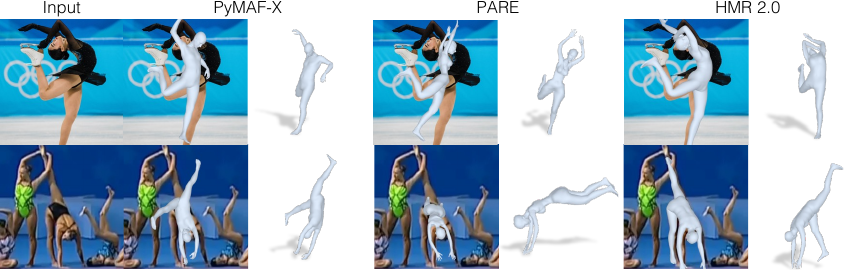}
    \caption{{\bf Qualitative comparison of state-of-the-art mesh recovery methods.} \approachName~returns more faithful reconstructions for unusual poses compared to the closest competitors, PyMAF-X~\cite{zhang2022pymaf} and PARE~\cite{kocabas2021pare}.}
    \label{fig:pose-comparison}
    \vspace{+0.3em}
\end{figure*}

\noindent
{\bf Versatility of PHALP$'$.}
With the modifications of PHALP$'$, we abandon the model-specific latent space of~\cite{rajasegaran2022tracking} and instead, we operate in the SMPL space, which is shared across most mesh recovery systems.
This makes PHALP$'$ more versatile and allows us to plug in different 3D pose estimators and compare them based on their performance on the downstream task of tracking.
We perform this comparison in \Cref{tab:tracking-ablation} where we use pose and location cues from state-of-the-art 3D pose estimators (while still using appearance from HMAR~\cite{rajasegaran2022tracking}).
We observe that \approachName~performs the best and PARE~\cite{kocabas2021pare}, HMAR~\cite{rajasegaran2022tracking}, and PyMAF-X~\cite{zhang2022pymaf} closely follow on the Posetrack dataset, with minor differences between them.
Note that tracking is often most susceptible to errors in predicted 3D locations with body pose having a smaller effect in performance~\cite{rajasegaran2022tracking}.
This means that good tracking performance can indicate robustness to occlusions, so it is helpful to consider this metric, but it is less helpful to distinguish fine-grained differences in pose.
As a result, the competitive results of PARE~\cite{kocabas2021pare}, HMAR~\cite{rajasegaran2022tracking}, and PyMAF-X~\cite{zhang2022pymaf} indicate that they handle occlusions gracefully, but their pose estimation might still be less accurate (as observed from \Cref{tab:2d_pck}).
See also \Cref{fig:pose-comparison} and SupMat for more qualitative comparisons.

\noindent
{\bf 4DHumans.}  
\Cref{tab:tracking} evaluates tracking performance of our complete system, \systemName, on the PoseTrack dataset. Using the same bounding box detector as~\cite{rajasegaran2021tracking, rajasegaran2022tracking}, \systemName~outperforms existing approaches on all metrics, improving ID Switches by 22\%. Using the improved ViTDet detector~\cite{li2022exploring} can improve performance further.
As a by-product of our temporal prediction model (\Cref{fig:phalp_pose_transformer}), we can perform amodal completion and attribute a pose to missing detections.
We show examples of this in the SupMat.

\addtolength{\tabcolsep}{-3pt}
\begin{table}[t]
\begin{center}
\begin{tabular}{l l c c c c }
\toprule[0.4mm]
\multirow{2}{*}{Tracker} &  \multirow{2}{*}{Pose Engine} &  \multicolumn{4}{c}{Posetrack} \\
                          &  &  HOTA$\uparrow$ & IDs$\downarrow$  & MOTA$\uparrow$ & IDF1$\uparrow$ \\ 
\midrule
\multirow{6}{*}{PHALP$'$} 
         & PARE~\cite{kocabas2021pare}             & \tbest{53.6}& 510        & \best{59.4}  & 76.8 \\
         & PyMAF-X~\cite{zhang2022pymaf}           & \sbest{53.7}& \sbest{472}& 59.2         & \tbest{76.9}\\
         & CLIFF~\cite{li2022cliff}                & 53.5        & 551        & 58.7         & 76.5 \\
         & PyMAF~\cite{zhang2021pymaf}             & 53.0        & 623        & 58.6         & 76.1 \\
         & HMAR~\cite{rajasegaran2022tracking}     & \tbest{53.6}& \tbest{482}& \tbest{59.3} & \sbest{77.1}\\
         & \approachName                           & \best{54.1} & \best{456} & \best{59.4}  & \best{77.4} \\ %
\bottomrule[0.4mm]
\end{tabular}
\end{center}
\caption{{\bf Tracking with different 3D pose estimators.} 
With the modifications of PHALP$'$, we have a versatile tracker that allows different 3D pose estimators to be plugged into it.
\approachName, PARE, and PyMAF-X perform the best in this setting.}
\vspace{0.2cm}
\label{tab:tracking-ablation}
\end{table}
\addtolength{\tabcolsep}{3pt}

\addtolength{\tabcolsep}{-3pt}
\begin{table}[t]
\begin{center}
\begin{tabular}{l c c c c }
\toprule[0.4mm]
\multirow{2}{*}{Method} & \multicolumn{4}{c}{Posetrack} \\
                            & HOTA$\uparrow$  & IDs$\downarrow$ & MOTA$\uparrow$ & IDF1$\uparrow$ \\ 
\midrule
Trackformer~\cite{meinhardt2022trackformer}   & 46.7         & 1263       & 33.7       & 64.0       \\
Tracktor~\cite{bergmann2019tracking}          & 38.5         & 702        & 42.4       & 65.2        \\
AlphaPose~\cite{fang2017rmpe}                  & 37.6          & 2220       & 36.9       & 66.9      \\
Pose Flow~\cite{xiu2018pose}                    & 38.0          & 1047       & 15.4       & 64.2     \\
T3DP~\cite{rajasegaran2021tracking}             & \tbest{50.6}        & \tbest{655}& \tbest{55.8} & \tbest{73.4} \\
PHALP~\cite{rajasegaran2022tracking}            & \sbest{52.9}    & \sbest{541}& \sbest{58.9} & \sbest{76.4}     \\
\systemName                                     & \best{54.3}     & \best{421} & \best{59.8} & \best{77.9}    \\ \midrule
\systemName~+ ViTDet                           & 57.8     & 382 & 61.4 & 79.1     \\
\bottomrule[0.4mm]
\end{tabular}
\end{center}
\caption{{\bf Comparison of \systemName~with the state of the art on the Posetrack dataset.} \systemName~achieve state-of-the-art tracking performance for all metrics. Incorporating a better detection system~\cite{li2022exploring} leads to further performance improvements.}
\label{tab:tracking}
\end{table}
\addtolength{\tabcolsep}{3pt}

\begin{table}[t]
\begin{center}
\begin{tabular}{c l c c c c}
\toprule[0.4mm]
Action & Pose & \multirow{2}{*}{OM}  & \multirow{2}{*}{PI} & \multirow{2}{*}{PM} & \multirow{2}{*}{mAP} \\
Model & Engine & & & & \\ \midrule
\multirow{6}{*}{\cite{rajasegaran2023action}} & PyMAF~\cite{zhang2021pymaf} & 7.3 & 16.9 & 34.7 & 15.4 \\
& CLIFF~\cite{li2022cliff} & \tbest{9.2} & 20.0 & 40.3 & 18.6 \\
& HMAR~\cite{rajasegaran2022tracking} & 8.7  & 20.1 & 40.3 & 18.3 \\
& PARE~\cite{kocabas2021pare} & \tbest{9.2} & \tbest{20.7} & \sbest{41.5} & \tbest{19.1} \\
& PyMAF-X~\cite{zhang2022pymaf} & \sbest{10.2} & \sbest{21.4} & \tbest{40.8} & \sbest{19.6} \\
& \approachName                 & \best{11.9} & \best{24.6} & \best{45.8} & \best{22.3} \\
\bottomrule[0.4mm]
\end{tabular}
\end{center}
\caption{\textbf{Action recognition results on the AVA dataset.} We benchmark different mesh recovery methods on the downstream task of pose-based action recognition. Here, \textit{OM} : Object Manipulation, \textit{PI} : Person Interactions, and \textit{PM} : Person Movement. 
}
\label{tab:action_pose_only}
\end{table}

\begin{figure*}
    \centering
    \includegraphics[width=\linewidth]{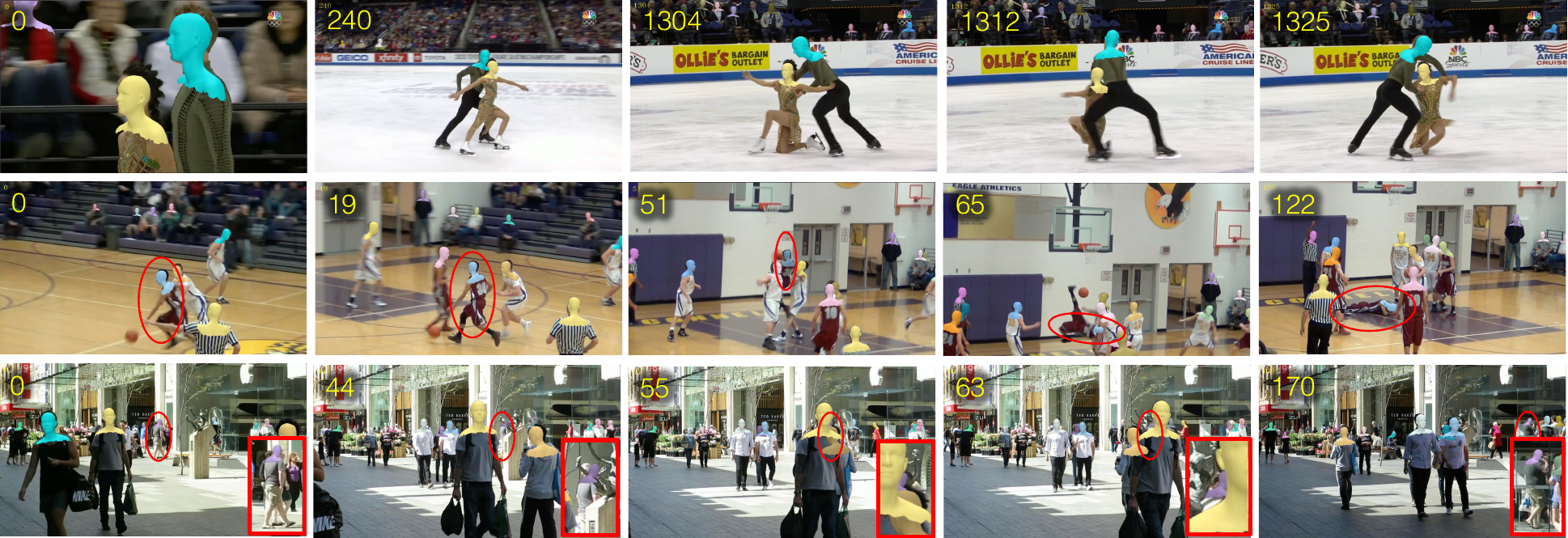}
    \caption{{\bf Qualitative tracking results of \systemName}. We use head masks (frame number is on the top left). First row: We track people skating on ice with challenging poses and heavy occlusions, in a minute long video without switching identities. Second row: The main person is tracked through multiple interactions with other players.
    Third row: The person of interest is tracked through long occlusions.}
    \label{fig:tracking-qual}
    \vspace{+0.3em}
\end{figure*}

\subsection{Action Recognition}
\label{sub:action}
\noindent
{\bf Evaluation setting.} The approach of~\cite{rajasegaran2023action} is the state of the art for action recognition in videos. Given a video as input, the authors propose using per-frame 3D pose and location estimates (using off-the-shelf HMR models~\cite{rajasegaran2022tracking}) as an additional feature for predicting action labels. They also show results for a ``pose-only'' baseline that predicts action labels using only 3D pose and location estimates. We use this setting to compare our model with baselines on the downstream task of action recognition on the AVA dataset~\cite{gu2018ava}.
In~\cite{rajasegaran2023action}, the authors train a transformer that takes SMPL poses as input and predicts action labels. Following their setup, %
we train a separate action classification transformer for each baseline. 

\noindent
{\bf Comparisons.}
Comparing results in \Cref{tab:action_pose_only}, we observe that \approachName~outperforms baselines on the different class categories (OM, PI, PM) and overall. It achieves an mAP of 22.3 on the AVA test set, which is 14\% better than the second-best baseline. Since accurate action recognition from poses needs fine-grained pose estimation, this is strong evidence that \approachName~predicts more accurate poses than existing approaches. 
In fact, when combined with appearance features, \cite{rajasegaran2023action} shows that \approachName~achieves the state of the art of 42.3 mAP on AVA action recognition, which is 7\% better than the second-best of 39.5 mAP.

\subsection{\approachName~Model Design}
\label{sub:design}

\begin{figure}[t]
\begin{center}
   \includegraphics[width=0.92\linewidth]{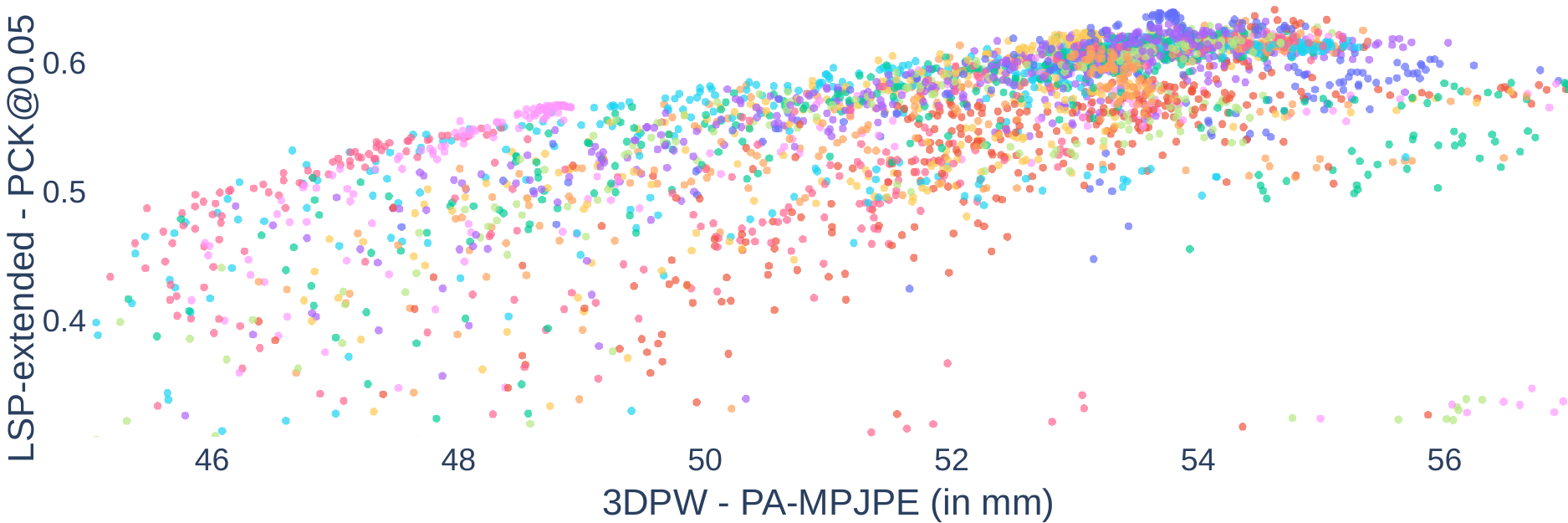}
\end{center}
   \caption{
   \textbf{Extensive model search.}
With each dot, we visualize the performance of a checkpoint when evaluated on 3DPW and LSP-Extended.
Colors indicate different runs.
We explore more than 100 settings, and visualize $\sim$100 checkpoints from each run.
   }
\label{fig:rebut:tradeoff}
\end{figure}

In the process of developing~\approachName, we investigated a series of design decisions.
\Cref{fig:rebut:tradeoff}~briefly illustrates this exploration.
We experimented with over 100 settings and we visualize the performance of 100 checkpoints for each run.
For the visualization, we use the performance of each checkpoint on the 3DPW and the LSP-Extended dataset.

Our investigation focused on some specific aspects of the model, which we document here as a series of ``lessons learnt'' for future research.
In the following paragraphs, we will regularly refer to \Cref{tab:rebut:metrics}, which evaluates these aspects on 3D and 2D metrics, using the 3DPW, Human3.6M, and LSP-Extended datasets.

\noindent
{\bf Effect of backbone.}
Unlike the majority of the previous work on Human Mesh Recovery that uses a ResNet backbone, our \approachName~method relies on a ViT backbone.
For a direct comparison of the effect of the backbone,
Model \texttt{B1} implements HMR with a ResNet-50 backbone and an MLP-based head implementing IEF (Iterative Error Feedback~\cite{carreira2016human,kanazawa2018end}).
In contrast, Model \texttt{B2} uses a transformer backbone (ViT-H) while keeping the other design decisions the same.
By updating the backbone, we observe a significant improvement across the 3D and 2D metrics, which justifies the ``transformerization'' step.

\noindent
{\bf Effect of training data.}
Besides the architecture, we also investigated the effect of training data.
Model \texttt{D1} trains on the typical datasets (H3.6M, MPII, COCO, MPI-INF) that most of the previous works are leveraging.
In comparison, model \texttt{D2} adds AVA in the training set, following~\cite{gu2018ava}.
Eventually, we also train using AI-Challenger and Insta-Variety (model \texttt{B2}), to further expand the training set.
As we can see, adding more training data leads to improvements across the board for the reported metrics, but the benefit is smaller compared to the backbone update.

\noindent
{\bf ViT pretraining.}
Another factor that had significant effect on the performance of our model was the pretraining of the ViT backbone.
Starting with randomly initialized weights (model \texttt{P1}) results in slow convergence and poor performance.
Results improve if our backbone is pretrained with MAE~\cite{he2022masked} on Imagenet (\texttt{P2}).
Eventually, our model of choice ({\approachName}b), which is first pretrained with MAE on ImageNet and then on the task of 2D keypoint prediction~\cite{xu2022vitpose}, achieves the best performance.

\noindent
{\bf SMPL head.}
We also investigate the effect of the architecture for the head that predicts the SMPL parameters.
Our proposed transformer decoder ({\approachName}b) improves performance when it comes to the image-model alignment (\ie, 2D metrics) compared to the traditional MLP-based head with IEF steps (\texttt{B2}).

\noindent
{\bf Dataset quality.}
Similar to previous work, \eg, [35],
it was crucial to keep the quality of the training data high; we filter out low quality pseudo-ground truth fits (high fitting error) and prune images with low-confidence 2D detections.

\addtolength{\tabcolsep}{-5pt}
\begin{table}[t]
  \centering
  \scriptsize
{
    \begin{tabular}{cl|cc|cc|cc}
    \toprule
    \multicolumn{2}{c|}{\multirow{2}[4]{*}{Models}} & \multicolumn{2}{c|}{3DPW} & \multicolumn{2}{c|}{Human3.6M} & \multicolumn{2}{c}{LSP-Extended} \\
\cmidrule{3-8}    \multicolumn{2}{c|}{} & MPJPE & PA-MPJPE & MPJPE & PA-MPJPE & PCK@0.05 & PCK@0.1 \\
    \midrule
          & {\approachName}b & 81.3     & 54.3  & 50.0     & 32.4   & 0.53	  & 0.82 \\
    \midrule
          & \texttt{B1} & 85.2     & 56.8  & 58.9     & 41.4   & 0.35	  & 0.66 \\
          & \texttt{B2} & 79.7     & 53.4  & 51.4     & 34.4   & 0.48	  & 0.81 \\
    \midrule
          & \texttt{D1} & 84.1     & 54.8  & 54.5     & 35.1   & 0.45	  & 0.79 \\
          & \texttt{D2} & 80.2     & 53.3  & 52.4     & 34.9   & 0.46	  & 0.79 \\
    \midrule
          & \texttt{P1} & 98.9   &	61.7	  & 89.9	  & 58.7	  & 0.24	  & 0.52 \\
          & \texttt{P2} & 82.7   &	55.6	  & 49.3	  & 32.4	  & 0.52	  & 0.81 \\
    \bottomrule
    \end{tabular}%
    }
    \caption{\textbf{Ablations}: Evaluation for different model designs on the 3DPW, Human3.6M, and LSP-Extended datasets.
    }
  \label{tab:rebut:metrics}%
\end{table}%
\addtolength{\tabcolsep}{5pt}

\section{Conclusion}
We study the problem of reconstructing and tracking humans from images and video.
First, we propose \approachName, a fully ``transformerized'' version of a network for the problem of Human Mesh Recovery~\cite{kanazawa2018end}.
\approachName~achieves strong performance on the usual 2D/3D pose metrics,
while also acting as the backbone for our improved video tracker.
The full system, \systemName, jointly reconstructs and tracks people in video and achieves state-of-the-art results for tracking.
To further illustrate the benefit of our 3D pose estimator, \approachName, we apply it to the task of action recognition, where we demonstrate strong improvements upon previous pose-based baselines.

Our work pushes the boundary of the videos that can be analyzed with techniques for 3D human reconstruction.
At the same time, the improved results also demonstrate the type of limitations that need to be addressed in the future.
For example, the use of the SMPL model~\cite{loper2015smpl} creates certain limitations, and leveraging improved models would allow us to model hand pose and facial expressions~\cite{pavlakos2019expressive}, or even capture greater age variation, \eg, infants~\cite{hesse2019learning} and kids~\cite{patel2021agora, sun2022putting}.
Moreover, since we consider each person independently, our reconstructions are less successful at capturing the fine-grained nature of people in close proximity, \eg, contact~\cite{fieraru2020three,muller2023generative}.
Besides this, our reconstructions ``live'' in the camera frame, so for proper understanding of the action in a video, we need to consider everyone in a common world coordinate frame, by reasoning about the camera motion too~\cite{pavlakos2022one,ye2023decoupling,yuan2022glamr}.
Finally, lower input resolution can affect the quality of our reconstructions, which could be addressed by resolution augmentations~\cite{xu20213d}.

\paragraph{Acknowledgements} We thank members of the BAIR community for helpful discussions and StabilityAI for their generous compute grant. This work
was supported by BAIR/BDD sponsors, ONR MURI (N00014-21-1-2801), and the DARPA MCS program.

\clearpage

{\small
\bibliographystyle{iccv2023AuthorKit/ieee_fullname}
\bibliography{egbib.bib}
}

\clearpage


\section*{Supplementary material}

# Supplementary Material for:
# "Humans in 4D: Reconstructing and Tracking Humans with Transformers"

Shubham Goel    Georgios Pavlakos    Jathushan Rajasegaran    Angjoo Kanazawa*    Jitendra Malik*

{shubham-goel, pavlakos, jathushan, kanazawa}@berkeley.edu, malik@eecs.berkeley.edu

University of California, Berkeley

We provide more details about HMR 2.0, *i.e.*, the architecture we use (Section S.1), the data (Section S.2) and the training pipeline (Section S.3). Furthermore, we describe the aspect of pose prediction (Section S.4) and we discuss the metrics we use for evaluation (Section S.5). Then, we discuss the experimental settings for tracking (Section S.6), and action recognition (Section S.7). Finally, we provide additional qualitative results (Section S.8).

## S.1. HMR 2.0 architecture details

The architecture of our HMR 2.0 model is based on a ViT image encoder and a transformer decoder. We use a ViT-H/16 ("huge") pre-trained on the task of 2D keypoint localization [25]. It has 50 transformer layers, takes a $256 \times 192$ sized image as input, and outputs $16 \times 12$ image tokens, each of dimension 1280. Our transformer decoder is a standard transformer decoder architecture [23] with 6 layers, each containing multi-head self-attention, multi-head cross-attention, and feed-forward blocks, with layer normalization [2]. It has a 2048 hidden dimension, 8 (64-dim) heads for self- and cross-attention, and a hidden dimension of 1024 in the feed-forward MLP block. It operates on a single learnable 2048-dimensional SMPL query token as input and cross-attends to the $16 \times 12$ image tokens. Finally, a linear readout on the output token from the transformer decoder gives pose $\theta$, shape $\beta$, and camera $\pi$.

## S.2. Data details

In our training, we adopt the training data conventions of previous works [10], using images from Human3.6M [4], COCO [13], MPII [1] and MPI-INF-3DHP [18]. This forms the training set for the version we refer to as HMR 2.0a in the main manuscript. For the eventual HMR 2.0b version, we additionally generate pseudo-ground truth SMPL [14] fits for images from AVA [3], InstaVariety [6] and AI Challenger [24]. Since AVA and InstaVariety include videos, we collect frames by sampling at 1fps and 5fps respectively. For pseudo-ground truth generation, we use ViTDet [11] for bounding box detection and ViTPose [25] for keypoint detection, while fitting happens using ProHMR [10]. We discard detections with very few 2D detected keypoints (less than five) and low detection confidence (threshold 0.5). We also discard fits with unnatural body shapes (*i.e.*, body shape parameters outside $[-3, 3]$), unnatural body poses (computed using a per-joint histogram of poses on AMASS [17]), and large fitting errors (*i.e.*, which indicates that the reconstruction was not successful). For training our HMR 2.0b model, we sample with different probabilities from each dataset, *i.e.*, Human3.6M: 0.1, MPII: 0.1, MPI-INF-3DHP: 0.1, AVA: 0.15, AI Challenger: 0.15, InstaVariety: 0.2, COCO: 0.2.

## S.3. Training details

We train our main model using 8 A100 GPUs with an effective batch size of $8 \times 48 = 384$. We use an AdamW optimizer [15] with a learning rate of 1e-5, $\beta_1 = 0.9$, $\beta_2 = 0.999$, and a weight decay of 1e-4. Training lasts for 1M iterations, which takes roughly six days. For our main model HMR 2.0b, we train the network end-to-end. However, for the HMR 2.0a variant, the ViT encoder remains frozen, allowing a larger effective batch size of $8 \times 512 = 4096$, learning rate of 1e-4, and fewer training iterations of 100K (*i.e.*, roughly equivalent number of epochs).

While training, we weigh the different losses. $\mathcal{L}_{\texttt{kp3D}}$, $\mathcal{L}_{\texttt{kp2D}}$, and $\mathcal{L}_{\texttt{adv}}$ have weights 0.05, 0.01, and 0.0005 respectively. The terms within $\mathcal{L}_{\texttt{smpl}}$ are also weighed differently, the $\theta$ and $\beta$ terms weigh 0.001 and 0.0005 respectively.

## S.4. Pose prediction

For the pose prediction model, we train a vanilla transformer model [23] from the tracklets obtained by [19]. Each tracklet at every time instance contains 3D pose and 3D location information, where the pose is parameterized by the SMPL model [14] and the location is represented as the translation in the camera frame. The transformer has 6 layers and 8 self-attention heads with a hidden dimension of 256. Each output token regresses the 3D pose and 3D loca-

tion of the person at the specified time-step. We train this model by randomly masking input pose tokens and applying the loss on the masked tokens. During inference, to predict a future 3D pose, we query the model by reading out from a future time-step, using a learned mask-token as input to that time-step. Similarly for amodal completion, we replace the missing detections with the learned mask-token and read out from the output at the corresponding time-step. The model is trained with a batch size of 64 sequences and a sequence length of 128 tokens. We use the AdamW optimizer [15] with a learning rate of 0.001 and $\beta_1 = 0.9, \beta_2 = 0.95$.

## S.5. Metrics

For our evaluation, we use the metrics that are common in the literature:
**3D Pose**: We follow [5] and we use MPJPE and PA-MPJPE. MPJPE refers to Mean Per Joint Position Error and it is the average L2 error across all joint, after aligning with the root node. PA-MPJPE is similar but is computed after aligning the predicted pose with the ground-truth pose using Procrustes Alignment.
**2D Pose**: We use PCK as defined in [26]. This is the Percentage of Correctly localized Keypoints, where a keypoint is considered as correctly localized if its L2 distance from the ground-truth keypoint is less than a threshold $t$. We report results using different thresholds (@0.05 and @0.1 of image size).
**Tracking**: Following [20, 21], we use standard tracking metrics. This includes ID switches (IDs), MOTA [7], IDF1 [22], and HOTA [16].
**Action Recognition**: We report results using mAP metrics as defined in the AVA dataset [3]. We further provided a more fine-grained analysis reporting results on different action categories: actions that involve Object Manipulation (OM), actions that involve Person Interactions (PI), and actions that involve Person Movement (PM). The results in these categories are also reported using mAP.

## S.6. Tracking with PHALP′

In the main manuscript, we compare different human mesh recovery systems on the downstream problem of tracking (Table 3 of the main manuscript). For this, we modify the PHALP approach [21], so that pose distance is computed on the SMPL space that all the models share. To make this comparison fair, we keep other variables similar to the original PHALP (*e.g.*, same appearance embedding). Note that this comparison is generous to baselines that do not model appearance themselves. Eventually, our final 4DHumans system uses a sampling-based appearance head and our new pose prediction, which lead to the state-of-the-art performance for tracking on PoseTrack (Table 4 of the main manuscript). To model appearance, we texture visible points on the mesh by projecting them onto the input image and sampling color from the corresponding pixels.

## S.7. Action recognition

As an alternative way to assess the quality of 3D human reconstruction, we evaluate various human mesh recovery systems on the downstream task of action recognition on AVA (please refer to [19] for more details on the task definition). More specifically, we take the tracklets from [19], which were generated by running PHALP [21] on the Kinetics [8] and AVA [3] datasets. Then, we replace the poses from various human mesh recovery models (*i.e.*, PyMAF [28], PyMAF-X [27], PARE [9], CLIFF [12], HMAR [21], HMR 2.0) and evaluate their performance on the action recognition task. In this pose-only setting, the action recognition model has access only to the 3D poses (in the SMPL format) and 3D location and is trained to predict the action of each person. For a fair comparison and to achieve the best performance for each 3D pose regressor, we retrain the action recognition model specifically for each 3D pose method.

## S.8. Additional qualitative results

We have already provided a lot of qualitative results of HMR 2.0, both in the main manuscript and in videos on the project webpage. Here, we provide additional results, including comparisons with our closest competitors (Figure S.1), and a demonstration of our results in a variety of challenging cases, including successes (Figure S.2) and failure cases (Figure S.3).

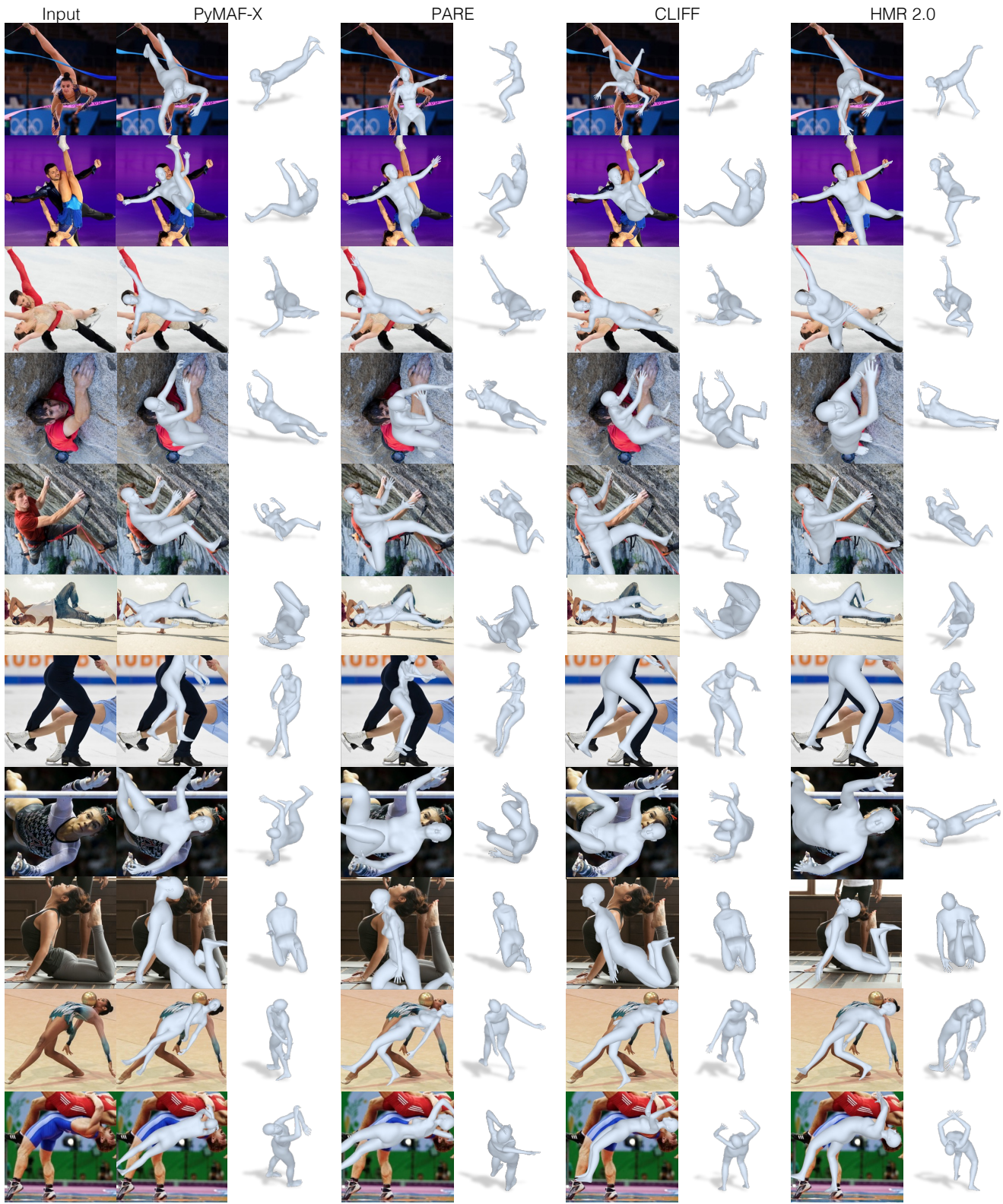

Figure S.1: **Qualitative comparison of our approach with state-of-the-art methods.** We compare HMR 2.0 with our closest competitors, PyMAF-X [27], PARE [9] and CLIFF [12]. For each example, we show the input image, and results from each method (including the frontal and a side view). HMR 2.0 is significantly more robust in a variety of settings, including images with unusual poses, unusual viewpoints and heavy person-person overlap.

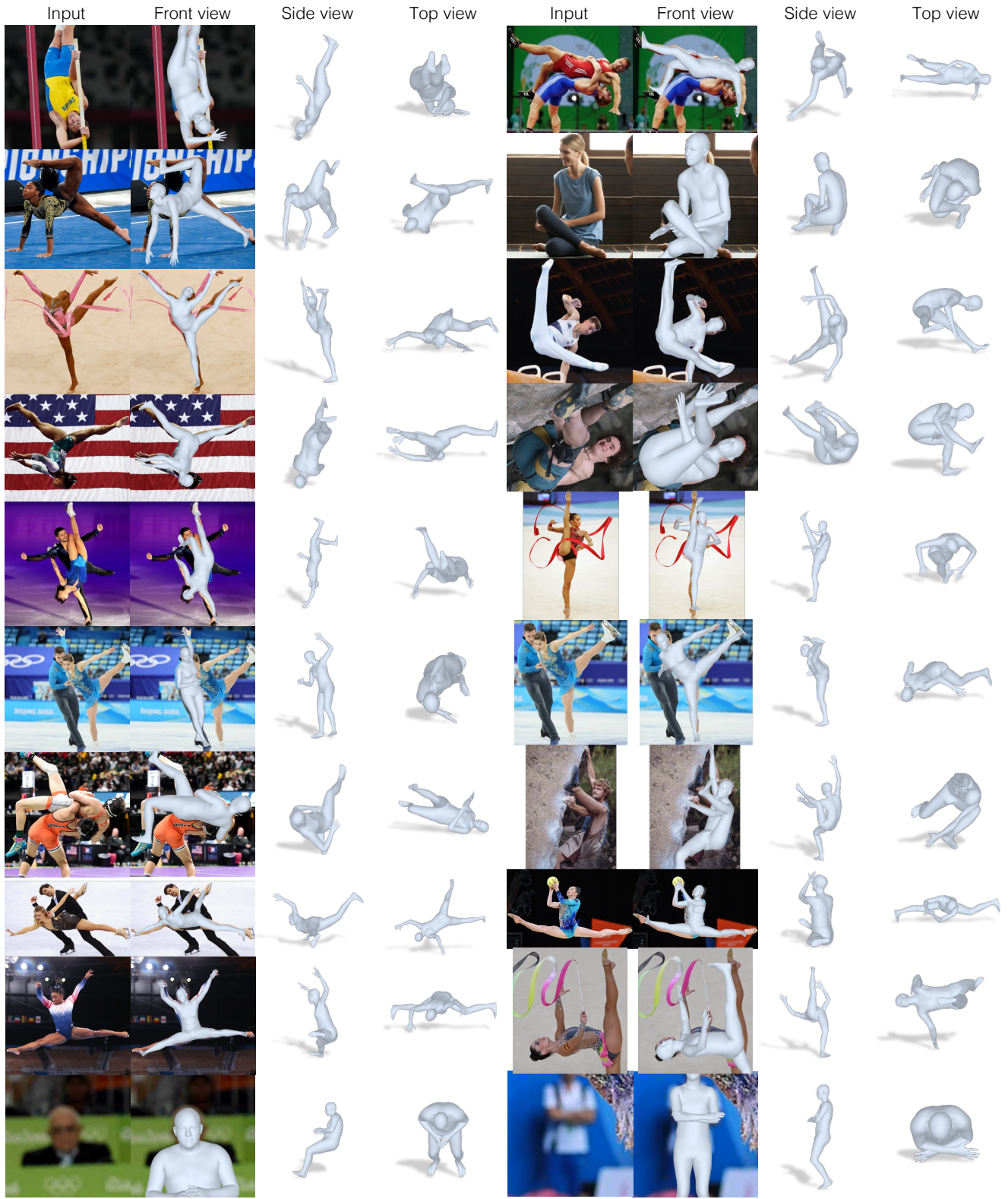

Figure S.2: **Qualitative results of our approach on challenging examples.** For each example we show the input image, the reconstruction overlay, a side view and the top view. The examples include unusual poses, unusual viewpoints, people in close interaction, extreme truncations and occlusions, as well as blurry images.

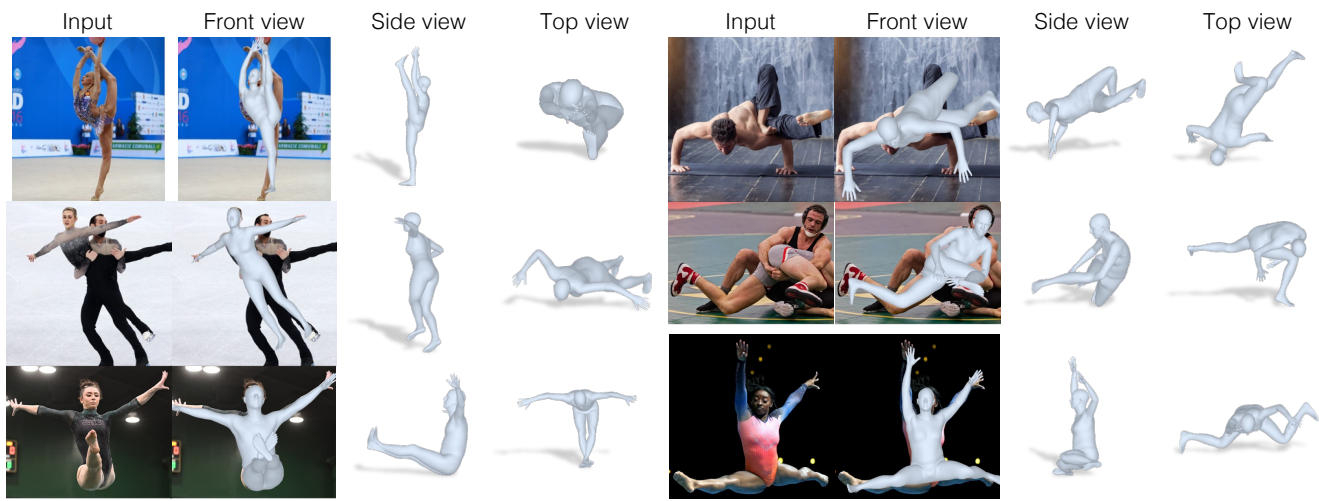

Figure S.3: **Failures of single frame 3D human reconstruction with HMR 2.0.** Despite the increased robustness of our method, we observe that HMR 2.0 occasionally recovers erroneous reconstructions in cases with very unusual articulation (first row), heavy person-person interaction (second row), and very challenging depth ordering for the different body parts (third row).



\end{document}

%% file: custom_includes.tex
\usepackage{xcolor}
\usepackage{colortbl}
\definecolor{turquoise}{cmyk}{0.65,0,0.1,0.3}
\definecolor{purple}{rgb}{0.65,0,0.65}
\definecolor{dark_green}{rgb}{0, 0.5, 0}
\definecolor{orange}{rgb}{0.8, 0.6, 0.2}
\definecolor{red}{rgb}{0.8, 0.2, 0.2}
\definecolor{blueish}{rgb}{0.0, 0.7, 1}
\definecolor{light_gray}{rgb}{0.7, 0.7, .7}
\definecolor{pink}{rgb}{1, 0, 1}
\definecolor{dark_red}{rgb}{0.5, 0, 0}

\definecolor{gold}{HTML}{EBC634}
\definecolor{silver}{HTML}{B4B4B4}
\definecolor{bronze}{HTML}{AD8A56}

\definecolor{light_silver}{HTML}{EBEBEC}
\definecolor{light_gold}{HTML}{FFF080}
\definecolor{light_bronze}{HTML}{F0D9C1}

\definecolor{rank_red}{HTML}{FFB3B3}
\definecolor{rank_orange}{HTML}{FFD9B3}
\definecolor{rank_yellow}{HTML}{FFFFB3}

\definecolor{green1}{HTML}{55a630}
\definecolor{green2}{HTML}{c1fba4}
\definecolor{green3}{HTML}{FFFFB3}

\usepackage{pifont}
\usepackage{xcolor}

\def\real{\mathbb{R}}
\def\loss{\mathcal{L}}

\usepackage{bbm}

\newcommand{\approachName}{\mbox{HMR 2.0}}
\newcommand{\systemName}{\mbox{4DHumans}}

\newcommand{\best}[1]{{\cellcolor{rank_red} #1}}
\newcommand{\sbest}[1]{{\cellcolor{rank_orange} #1}}
\newcommand{\tbest}[1]{{\cellcolor{rank_yellow} #1}}

\definecolor{rblue}{RGB}{68, 114, 196}
\definecolor{rgreen}{RGB}{112, 173, 71}
\definecolor{rorange}{RGB}{237, 125, 49}